\PassOptionsToPackage{table}{xcolor}
\documentclass[10pt,twocolumn,letterpaper]{article}

\usepackage{wacv}              

\definecolor{wacvblue}{rgb}{0.21,0.49,0.74}
\definecolor{oursgray}{gray}{0.92}  
\definecolor{lightgrayrule}{gray}{0.85}
\usepackage{colortbl}
\usepackage[pagebackref,breaklinks,colorlinks,allcolors=wacvblue]{hyperref}
\usepackage{graphicx}
\usepackage[normalem]{ulem}
\usepackage{booktabs}
\usepackage{adjustbox}
\usepackage{multirow}
\usepackage[table]{xcolor}
\usepackage{makecell}
\usepackage[accsupp]{axessibility}

\hypersetup{
    colorlinks=true,
}

\def\wacvPaperID{1418} 
\def\confName{WACV}
\def\confYear{2026}

\title{GrowTAS: Progressive Expansion from Small to Large Subnets\\ for Efficient ViT Architecture Search}

\author{
Hyunju Lee \qquad
Youngmin Oh \qquad
Jeimin Jeon \qquad
Donghyeon Baek \qquad
Bumsub Ham\thanks{Corresponding author.}\\
Yonsei University\\
\url{https://cvlab.yonsei.ac.kr/projects/GrowTAS/}
}

\begin{document}
\maketitle



\begin{abstract}
    Transformer architecture search (TAS) aims to automatically discover efficient vision transformers (ViTs), reducing the need for manual design. Existing TAS methods typically train an over-parameterized network (\emph{i.e.}, a supernet) that encompasses all candidate architectures (\emph{i.e.}, subnets). However, subnets partially share weights within the supernet, which leads to interference that degrades the smaller subnets severely. We have found that well-trained small subnets can serve as a good foundation for training larger ones. Motivated by this, we propose a progressive training framework, dubbed GrowTAS, that begins with training small subnets and incorporates larger ones gradually. This enables reducing the interference and stabilizing training. We also introduce GrowTAS+ that fine-tunes a subset of weights only to further enhance the performance of large subnets. Extensive experiments on ImageNet and several transfer learning benchmarks, including CIFAR-10/100, Flowers, CARS, and INAT-19, demonstrate the effectiveness of our approach over current TAS methods. 
\end{abstract}    
\section{Introduction} \label{sec:intro}

\begin{figure}[t]
  \centering
  \small
  \includegraphics[width=1.0\linewidth]{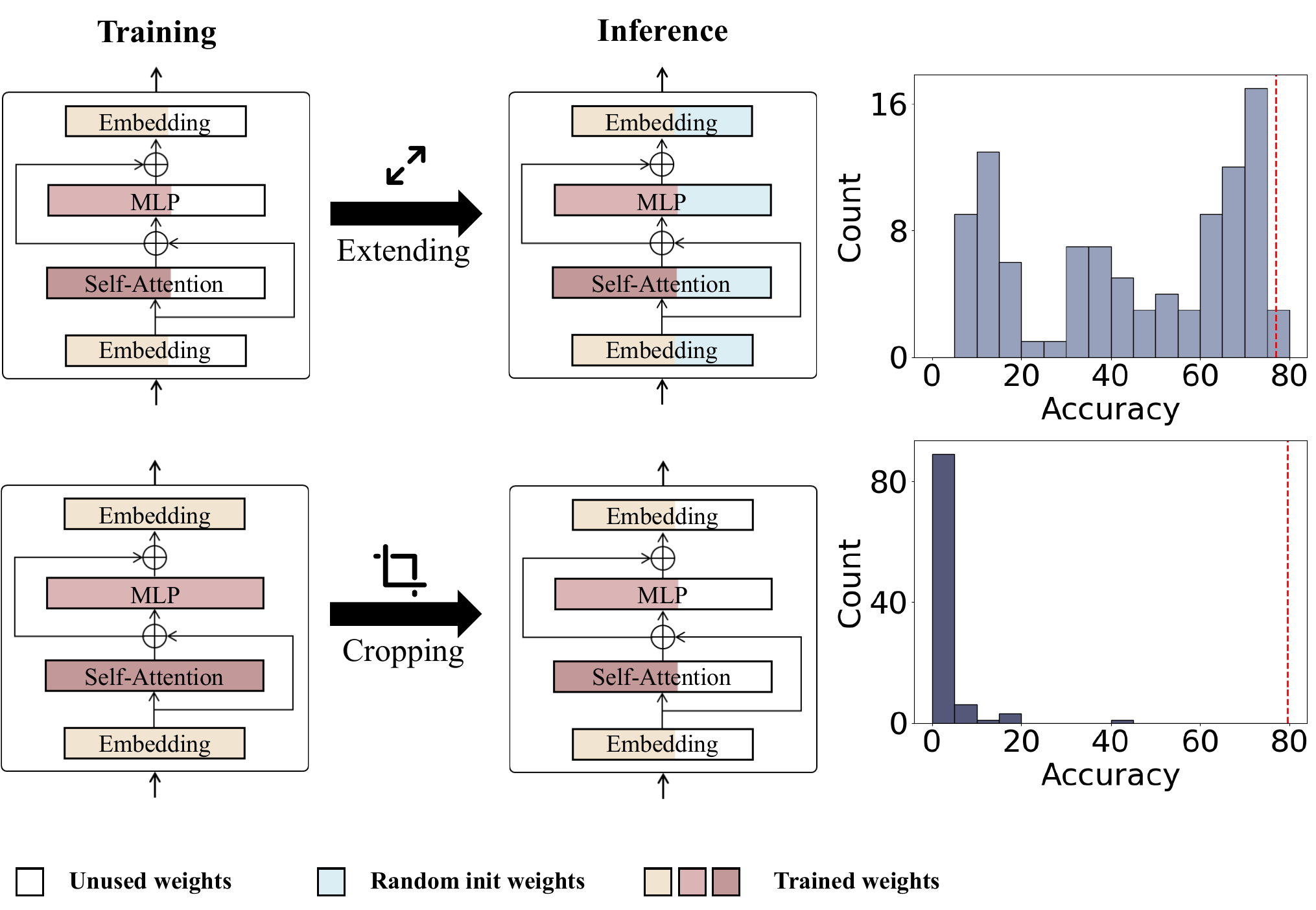}
  \caption{\textbf{Top}: We first train a small subnet and then evaluate 100 larger subnets derived from it without further training. To be specific, the large subnets are obtained by appending randomly initialized weights to those of the trained one. We can see that the large subnets achieve test accuracies comparable to the trained one (marked by the red dashed line) even with the random initialization. \textbf{Bottom}: We first train a large subnet and then evaluate 100 smaller subnets derived from it without further training. Specifically, we obtain the small variants by cropping a subset of weights from the trained subnet randomly. We can see that the small subnets fail to maintain the performance of the well-trained one.}
  \label{fig:variance_comparison}
\end{figure}

Vision Transformers (ViTs) have shown impressive results in various tasks such as image classification~\cite{dosovitskiy2020image}, object detection~\cite{carion2020end}, and semantic segmentation~\cite{strudel2021segmenter}. Compared to convolutional neural networks (CNNs)~\cite{he2016deep, tan2019efficientnet}, ViTs adopt a self-attention mechanism~\cite{vaswani2017attention} that enables long-range interactions between intermediate features, capturing structural information (\emph{e.g.}, spatial layouts) in images. However, designing ViT architectures requires lots of trial and error with human experts, which is time-consuming. Recently, transformer architecture search (TAS) has been proposed to automate the design process, finding optimal architectures for a specific hardware constraint (\emph{e.g.}, FLOPs). Following one-shot neural architecture search (NAS) approaches~\cite{guo2020single, pham2018efficient} for CNNs, the seminal work of~\cite{chen2021autoformer} proposes to optimize an over-parameterized transformer (\emph{i.e.}, a supernet) consisting of all candidate architectures (\emph{i.e.}, subnets) within a search space. After training, the supernet serves as a performance estimator for the various subnets, thereby avoiding training them from scratch. This enables identifying the top-performing transformer within the search space rapidly. However, the subnets sharing the same set of weights conflict with each other at training time, leading to suboptimal performance. Although several methods attempt to address the conflict by prioritizing promising subnets during training, they require either extra computational costs~\cite{liu2022focusformer} or re-training subnets whenever hardware constraints are changed~\cite{wang2023prenas}. In addition, these methods update a limited number of subnets only, which causes performance degradation, particularly for small subnets.

We propose in this paper a novel one-shot TAS framework, dubbed GrowTAS, that alleviates the interference between subnets during training, improving the search performance. In particular, we have found that (1) a well-trained small subnet can serve as a starting point for training larger subnets (Fig.~\ref{fig:variance_comparison} top). This is because the large subnets inherit a pre-trained set of weights from the smaller one, while adding randomly initialized weights. Since the influence of the random weights is marginal compared to that of the well-trained weights, the large subnets are likely to preserve the test accuracy of the trained small subnet. (2) smaller subnets generated from a well-trained large subnet fail to maintain the test accuracy (Fig.~\ref{fig:variance_comparison} bottom). This is because the small subnets derived from the larger one rely on cropping a subset of weights. This removes structural patterns (\emph{e.g.}, rows and columns) from the original set of weights, disrupting the learned dependencies. Motivated by the findings, we introduce a progressive training scheme for the supernet, where it trains small subnets initially and incorporates larger ones in a gradual manner. This prevents the small subnets from being disturbed by the larger ones at the early phase of training, while the well-trained subnets serve as good starting points for training larger ones. Although GrowTAS boosts search performance, especially for small subnets, the improvement for large subnets is limited in that the newly initialized set of weights are updated relatively little. To address this, we propose GrowTAS+ that fine-tunes the newly added weights only, while freezing the remaining ones. This allows us to further enhance the performance of large subnets, while preserving the performance of small subnets. Experimental results show that our approach achieves state-of-the-art performance on ImageNet~\cite{deng2009imagenet}, demonstrating its effectiveness. We also evaluate our method on CIFAR-10/100~\cite{krizhevsky2009learning}, Flowers~\cite{nilsback2008automated}, CARS~\cite{krause2013d}, and iNaturalist 2019~\cite{van2018inaturalist}, validating its transferability. Our contributions can be summarized as follows:
\begin{itemize}
  \item We perform an in-depth analysis to show that well-trained small subnets can serve as a good foundation for training larger ones, and introduce a novel one-shot TAS framework, dubbed GrowTAS, that trains subnets in a progressive manner to improve search performance.
  \item We present a simple yet effective strategy that enhances the performance of large subnets without sacrificing the performance of small ones.
  \item We provide extensive experiments on standard benchmarks~\cite{deng2009imagenet,krizhevsky2009learning,nilsback2008automated,krause2013d,van2018inaturalist}, demonstrating the effectiveness of our method.
\end{itemize}



\section{Related work} \label{sec:related_work}
Many methods have been introduced to identify top-performing architectures under hardware constraints. In particular, they can be categorized into two groups: NAS and TAS. While NAS focuses on identifying optimal convolutional neural networks, TAS aims to search best-performing transformer architectures. In the following, we describe representative works pertinent to ours.

\subsection{NAS}
NAS~\cite{zoph2017nas,zoph2018learning,baker2017designing, guo2020spos, yu2020bignas, cai2020once, lee2024aznas,jeon2025subnet,oh2025efficient} aims to automatically design high-performing architectures under given constraints. Early NAS approaches~\cite{zoph2017nas,zoph2018learning,baker2017designing} adopt reinforcement learning~\cite{williams1992simple} that requires training neural networks repeatedly, which is computationally demanding. To reduce the training cost, one-shot NAS approaches~\cite{guo2020spos, yu2020bignas, cai2020once} propose to train a single supernet that consists of all candidate subnets within a search space. Since subnets partially share weights from the supernet, the supernet is capable of predicting the performance of different subnets after training. Specifically, SPOS~\cite{guo2020spos} samples one subnet from the supernet at each training step, and updates the corresponding weights of the supernet only. It however treats all subnets equally during training without considering the performance of each subnet. GreedyNAS~\cite{you2020greedynas} instead focuses on training a set of promising subnets. It evaluates several subnets iteratively on a validation set to identify potentially good ones. Although all the aforementioned methods reduce the computational cost, they are limited in that the searched architectures require re-training from scratch to achieve optimal performance. To remove the re-training stage, OFA~\cite{cai2020once} introduces a progressive shrinking strategy that first trains the largest subnet only and fine-tunes smaller subnets gradually. Rather than fine-tuning the smaller subnets, BigNAS~\cite{yu2020bignas} trains the smallest, largest, and randomly sampled subnets at each training step simultaneously. This enables better training subnets of different sizes, while avoiding the re-training and fine-tuning stages. Similarly, our approach considers the size of subnets during training, but differs in that it starts from small subnets rather than large ones. In particular, our approach prevents the small subnets from being hindered by the larger ones at the early phase of training, improving the search performance.


\subsection{TAS} 
A relatively few methods~\cite{chen2021autoformer, liu2022focusformer, zhang2023shiftnas, wang2023prenas} address TAS, typically adopting the concept of one-shot NAS approaches. AutoFormer~\cite{chen2021autoformer} proposes to use slimmable neural networks~\cite{yu2019slimmable} to build a supernet consisting of all candidate subnets. It however samples subnets uniformly during training, suffering from conflicts between subnets. To address this, PreNAS~\cite{wang2023prenas} measures the trainability of each subnet by using a pruning technique~\cite{lee2019snip}, and then trains a few subnets that are highly trainable. Although PreNAS reduces the interference between subnets, it requires selecting and training subnets when hardware constraints are changed. Recently, FocusFormer~\cite{liu2022focusformer} introduces a trainable sampler that learns to identify top-performing subnets at training time. This allows high-quality subnets to be sampled more frequently, but the sampler induces extra computational overheads. Rather than training a few subnets only as in~\cite{wang2023prenas,liu2022focusformer}, our approach trains all subnets in a progressive manner. Specifically, we train small subnets first, preventing them from being disturbed by larger ones. The trained subnets then serve as a good foundation for training larger ones, allowing to better train all subnets. This is different from current methods~\cite{wang2023prenas,liu2022focusformer} in that they focus on improving a limited number of subnets only, while the remaining subnets remain unsatisfactory.

\section{Method}

In this section, we describe a brief overview of one-shot TAS~\cite{chen2021autoformer} (Sec.~\ref{sec:supernet_training}), and present a detailed description of our approach (Sec.~\ref{sec:growtas}). We then describe an evolutionary search method~\cite{guo2020spos} to identify optimal transformer under hardware constraints (Sec.~\ref{sec:search}).


\begin{figure}[t]
    \centering
    \small
    \begin{subfigure}{1.0\linewidth}
      \centering
      \includegraphics[width=\linewidth]{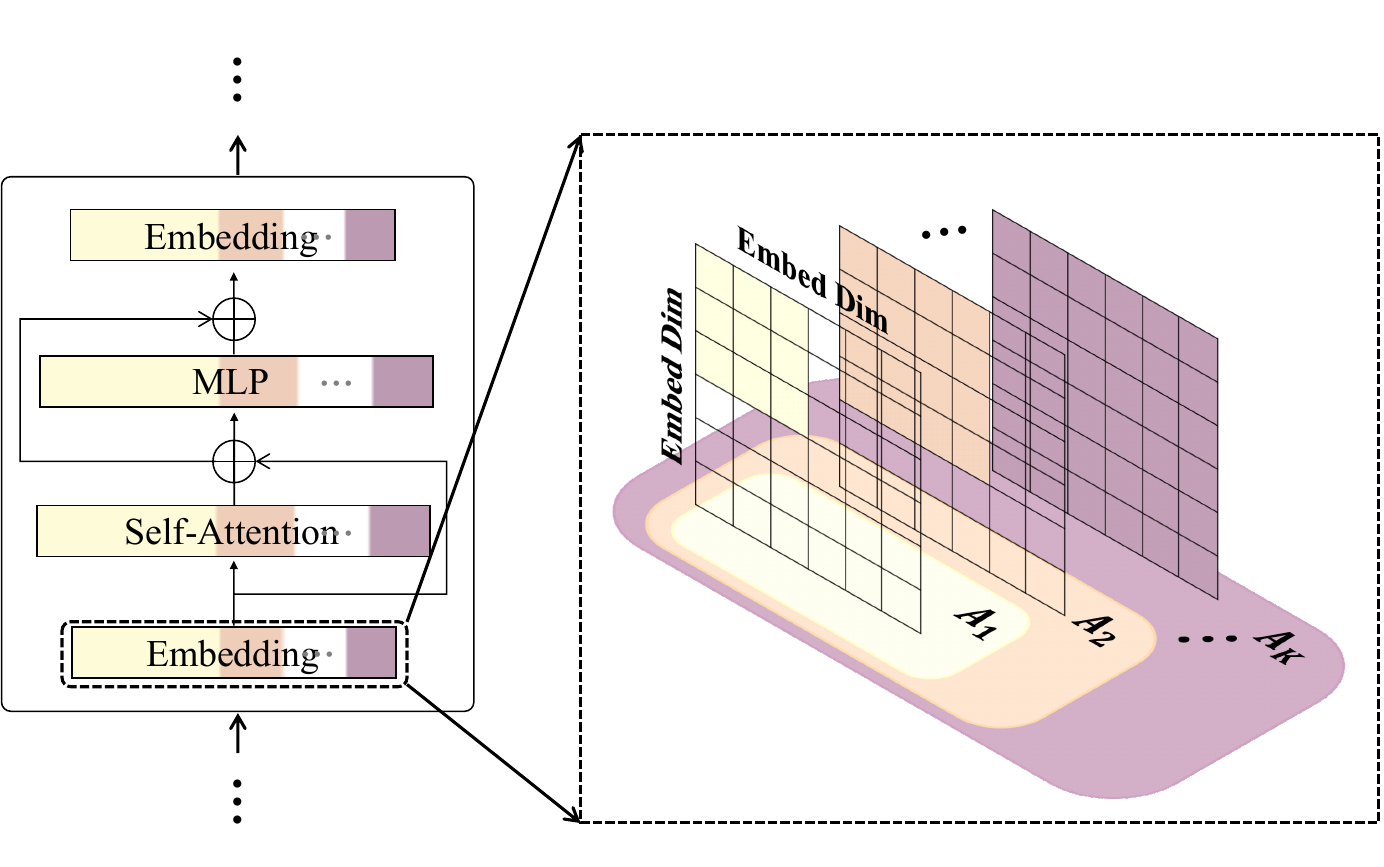}
    \end{subfigure}
	\caption{An illustration of the weight-sharing technique for TAS. We visualize a weight matrix of size Embed Dim $\times$ Embed Dim, where Embed Dim indicates the maximum number of channels in a search space. We use different colors to represent the corresponding region within the matrix for each subspace. We can see that the weights of a small subnet (\emph{e.g.}, from \( \mathcal{A}_1 \)) belong to those of larger ones (\emph{e.g.}, from \( \mathcal{A}_2 \) or \( \mathcal{A}_K \)).}
    \label{fig:weight_matrix}
\end{figure}

\begin{figure*}[t]
   \centering
   \small
   \includegraphics[width=1.0\linewidth]{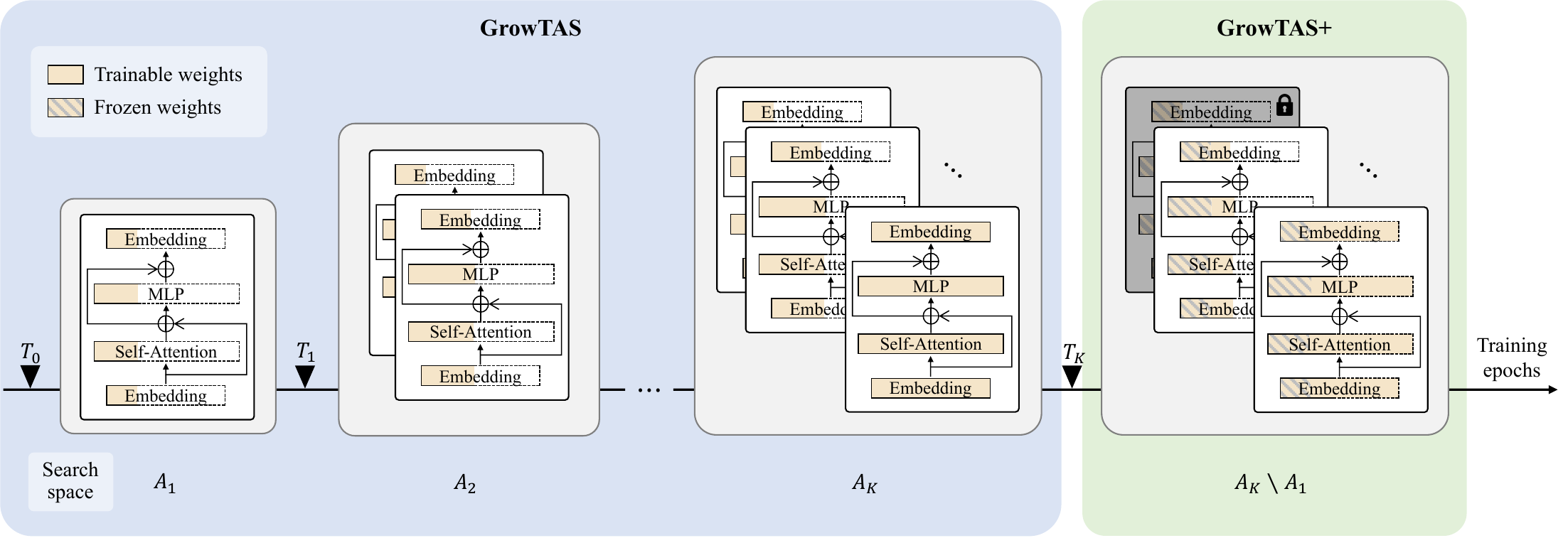}
   \caption{An overview of GrowTAS. We first divide the search space into a set of $K$ subspaces, and begin with training subnets from the smallest subspace \(\mathcal{A}_1\). This allows the small subnets to be trained with minimal interference, enabling better training them. We incorporate subnets from larger subspaces gradually, where large subnets can benefit from a set of well-trained weights from smaller ones. We introduce GrowTAS+ that fine-tunes a subset of weights only, while freezing the remaining weights. This allows us to improve the performance of large subnets without changing the well-trained weights of smaller subnets.}
   \label{fig:overview}
\end{figure*}

\subsection{One-shot TAS} \label{sec:supernet_training}
AutoFormer~\cite{chen2021autoformer} introduces a search space for transformers, where it consists of the embedding dimension, the expansion ratio, the number of attention heads, and the number of transformer blocks. Specifically, the number of heads and the expansion ratio differ across blocks, while the embedding dimension is the same for all blocks. Since training each transformer from scratch is computationally demanding, AutoFormer proposes to train a single supernet covering all candidate architectures (\emph{i.e.}, subnets) in the search space. At each training step, it uniformly samples one subnet from the search space and trains the corresponding weights of the supernet. Let us denote by $\alpha$ and $\mathcal{A}$ a certain subnet and the search space, respectively. We can then define a training objective for the supernet as follows:
\begin{equation}
    \mathcal{W}^* = \arg\min_{\mathcal{W}} \; \mathbb{E}_{\alpha \sim U(\mathcal{A})} \left[ \mathcal{L}_{\text{train}}(\alpha; \mathcal{W}) \right],	
    \label{eq:1}
\end{equation}
where we denote by $\mathcal{W}$ a set of trainable weights of the supernet and $\mathcal{L}_{\text{train}}(\alpha; \mathcal{W})$ indicates a training loss using the subnet $\alpha$. However, the uniform sampling strategy across the entire space, denoted by $U(\mathcal{A})$, treats all subnets equally during training, making small ones susceptible to interference. This is because the weights of small subnets are a subset of those of larger ones (See Fig.~\ref{fig:weight_matrix}).


%

\subsection{GrowTAS} \label{sec:growtas}
We propose a progressive training framework for TAS that trains subnets from small to large (See Fig.~\ref{fig:overview}). We first present an in-depth analysis to show that large subnets extended from a pre-trained small one are likely to preserve the performance even without further training. We then introduce our approach that first samples small subnets and incorporate larger ones gradually. We further propose a simple yet effective strategy that improves the performance of large subnets, while preserving the performance of small ones. Finally, we describe a detailed description of the evolutionary search algorithm used to identify optimal architectures from the trained supernet.

\begin{figure}[t]
    \centering
    \small
    \begin{subfigure}{1.0\linewidth}
      \centering
      \includegraphics[width=\linewidth]{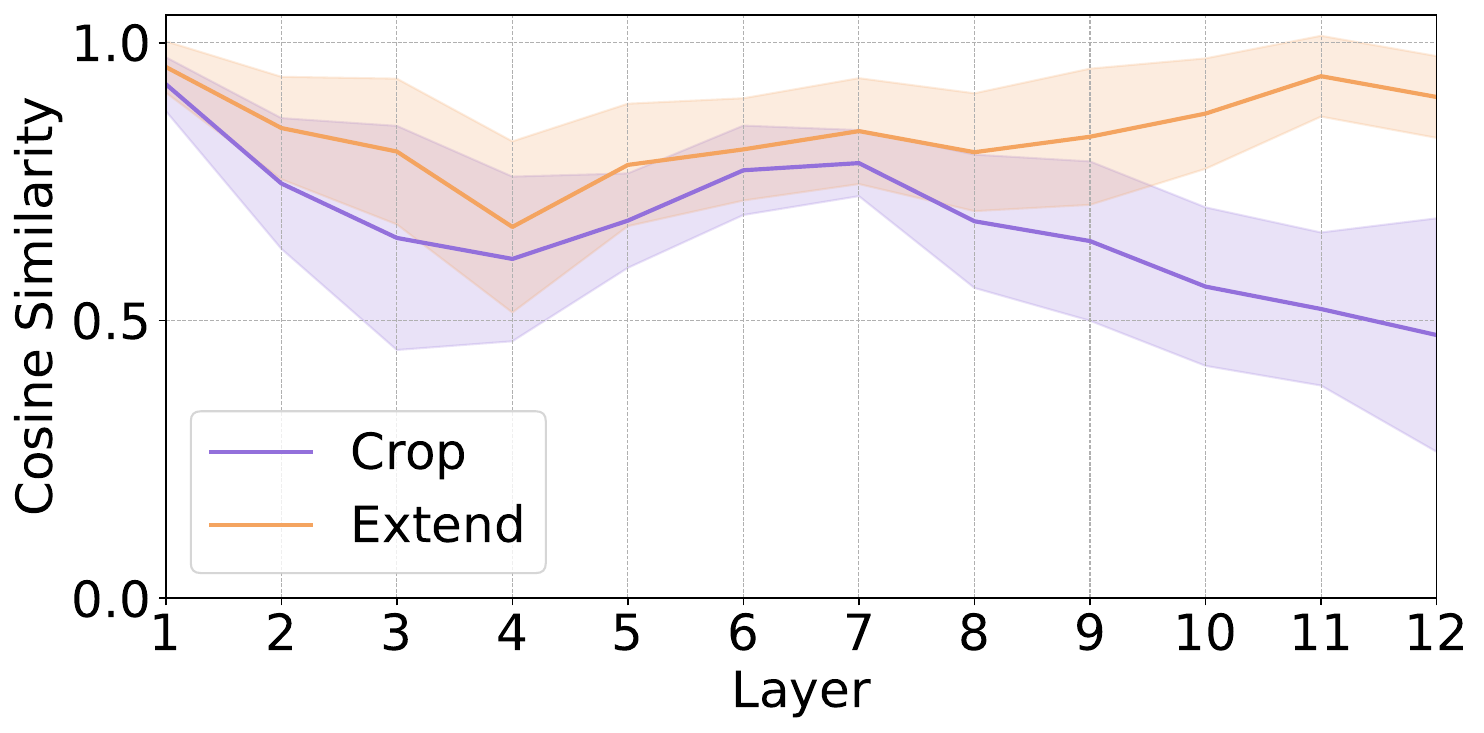}
    \end{subfigure}
    \caption{Comparison of layer-wise cosine similarities. We report the average cosine similarity along with standard deviations across all token pairs between the well-trained subnet and its variants. Please see text for details.}
    \label{fig:cos_sim_main}
\end{figure}

\paragraph{Analysis.} We have observed in Fig.~\ref{fig:variance_comparison} that small subnets are likely to serve as a good foundation for training larger ones. To understand the reason behind this, we measure feature similarity between the well-trained subnet and a subnet derived from it. Specifically, we show in Fig.~\ref{fig:cos_sim_main} the average cosine similarity along with standard deviations for token pairs at each transformer block. From this figure, we have two findings: (1) Large subnets produce feature representations similar to the well-trained subnet (marked by the orange line). This suggests that the influence of the newly initialized weights is weaker than that of the well-trained weights, especially during the early phase of training. (2) Small subnets, derived from the well-trained one, struggle to produce similar features (marked by the purple line), indicating that simply using a subset of well-trained weights induces a severe shift in the feature space. In particular, we can see that the cosine similarities of the small subnets are much lower than those of the large subnets in deeper layers. Motivated by these findings, we propose to begin with training small subnets only, preventing them from being disturbed by larger subnets. We then incorporate the large subnets gradually so that the smaller subnets serve as a good foundation.

\paragraph{Progressive Subnet Sampling.} To train subnets progressively from small to large, we first divide the search space into a set of subspaces based on the size of the subnets\footnote{We consider the embedding dimension and expansion ratio for each subnet, since they primarily determine its size.}. Specifically, let us suppose we have $K$ subspaces in total as follows:
\begin{equation}
	\mathcal{A}_1 \subset \mathcal{A}_2 \subset \dotsb \subset \mathcal{A}_k \subset \dotsb \subset \mathcal{A}_K.
\end{equation}
In particular, we set the entire space to the largest subspace, \emph{i.e.}, $\mathcal{A}_K = \mathcal{A}$. We visualize in Fig.~\ref{fig:weight_matrix} the relationship between subnets from different subspaces. We can see that smaller subnets are nested within larger ones in terms of both structure and weight coverage. For training the supernet, we introduce a progressive strategy that samples subnets from the smallest $\mathcal{A}_1$ to the largest subspace $\mathcal{A}_K$ gradually (Fig.~\ref{fig:overview} left). Concretely, we modify the training objective in Eq.~\eqref{eq:1} as follows:
\begin{equation}
    \mathcal{W}^* = \arg\min_{\mathcal{W}} \; \mathbb{E}_{\alpha \sim P(t)} \left[ \mathcal{L}_{\text{train}}(\alpha; \mathcal{W}) \right],	
\end{equation}
where $t$ indicates a current epoch and we denote by $P(t)$ our sampling strategy, defined as follows:
\begin{equation}
    P(t) = U(\mathcal{A}_k) ~~\text{for}~~ T_{k-1} \le t < T_{k}.
\end{equation}
We denote by $T_k$ each transition step, and set $T_0$ and $T_K$ to zero and the total number of training epochs, respectively. Note that $U(\mathcal{A}_k)$ represents the uniform sampling scheme across the corresponding subspace $\mathcal{A}_k$. At the early phase of training, our approach trains small subnets only, preventing them from being hindered by larger ones. This enables better training the small subnets. As the training process continues, the likelihood of sampling larger subnets also increases. This allows large subnets to benefit from the pre-trained ones.

\paragraph{GrowTAS+.} Although GrowTAS enhances the search performance, especially for small subnets, it produces a limited improvement for large subnets. This is because the frequency of updating the newly initialized set of weights for large subnets during training is relatively lower compared to the frequency of updating the set of weights shared with the smaller ones. To compensate this, we introduce a simple yet effective strategy that fine-tunes a subset of weights only (Fig.~\ref{fig:overview} right). Formally, we further optimize the supernet for a small number of training epochs as follows:
\begin{equation}
    \mathcal{\bar{W}} = \arg\min_{\mathcal{W}^*} \; \mathbb{E}_{\alpha \sim U(\bar{\mathcal{A}})} \left[ \mathcal{L}_{\text{train}}(\alpha; \mathcal{W}^*) \right],	
\end{equation}
where $\bar{\mathcal{A}} = \mathcal{A}_K \setminus \mathcal{A}_1$. In particular, we freeze the corresponding set of weights for subnets within the smallest subspace $\mathcal{A}_1$, while fine-tuning the remaining set of weights. This allows us to preserve the performance of the small subnets, while improving the performance of the large ones with a negligible overhead.


\subsection{Evolutionary Search} \label{sec:search}
After training the supernet, we aim to identify optimal architectures under hardware constraints from the entire search space $\mathcal{A}$. Following the common practice~\cite{chen2021autoformer,su2022vitas}, we construct  a validation set by subsampling 10,000 training images from ImageNet~\cite{deng2009imagenet} (100 per class). Since evaluating all possible subnets is prohibitively expensive, we perform an evolutionary search method~\cite{guo2020spos,goldberg1991comparative} on the validation set. Concretely, we obtain the optimal architecture \( \alpha^* \) as follows:
\begin{equation}
    \alpha^* = \arg\min_{\alpha \in \mathcal{A}} \mathcal{L}_{\text{val}}(\alpha; \mathcal{W}_\text{pret}),
\end{equation}
where we denote by \( \mathcal{L}_{\text{val}} \) a validation loss using the subnet $\alpha$ and \( \mathcal{W}_\text{pret} \) can be either $\mathcal{W}^*$ or $\mathcal{\bar{W}}$. We perform the evolutionary search over 20 generations with a population size of 50. In each generation, the top-10 architectures in terms of the validation accuracy are selected as parents to produce the next generation. New candidate architectures are generated via mutation and crossover operations, where we set the mutation probability to 0.2.
\section{Experiments}

In this section, we describe training details~(Sec.~\ref{sec:settings}) and present experimental results of our approach on standard TAS benchmarks~(Sec.~\ref{sec:results}). We then provide discussions to validate the effectiveness of each component of our method~(Sec.~\ref{sec:analysis}).

\subsection{Experimental settings}
\label{sec:settings}

\subsubsection{Search space and datasets}
We train and evaluate our method on three search spaces~\cite{chen2021autoformer}: AutoFormer-T, AutoFormer-S, and AutoFormer-B~(See Table~\ref{tab:searchspace}). To show the effectiveness of our method, we mainly use the ImageNet dataset~\cite{deng2009imagenet}. In addition, we fine-tune the searched subnets from ImageNet on CIFAR-10/100~\cite{krizhevsky2009learning}, Flowers~\cite{nilsback2008automated}, Cars~\cite{krause2013d}, and INAT-19~\cite{van2018inaturalist} to assess the transferability of our method.

\subsubsection{Implementation details}

\paragraph{GrowTAS.}
We implement our method using the PyTorch~\cite{paszke2019pytorch} framework and the timm~\cite{wightman2019timm} library. Following~\cite{chen2021autoformer, wang2023prenas}, we train all models using the AdamW optimizer, a mini-batch size of 128, and a cosine learning rate scheduler over 500 epochs. For the AutoFormer-T space, the learning rate decays from 1e\text{-}3 to 2e\text{-}5, while for AutoFormer-S and AutoFormer-B spaces, it decays from 1e\text{-}3 to 2e\text{-}5 and 2e\text{-}7, respectively. We apply standard data augmentations including RandAugment~\cite{cubuk2020randaugment}, Mixup~\cite{zhang2018mixup}, CutMix~\cite{yun2019cutmix}, label smoothing~\cite{szegedy2016rethinking, yuan2020revisiting}, and repeated augmentation~\cite{berman2019multigrain,hoffer2020augment}. All experiments on AutoFormer-T and AutoFormer-S are conducted with 8 NVIDIA A10 GPUs, while AutoFormer-B is conducted on 8 NVIDIA A6000 GPUs. We set state transition points $T_0$ and $T_K$ to 0 and 500, respectively, where the number of subspaces $K$ is set to 2 for all models. The first transition point $T_1$ is set to 250, 225, and 200 for AutoFormer-T, AutoFormer-S, and AutoFormer-B spaces, respectively.

\paragraph{GrowTAS+.}
We fine-tune the supernet trained with GrowTAS for a small number of epochs. Specifically, on AutoFormer-T and AutoFormer-S spaces, we fine-tune the supernet for 20 epochs with an initial learning rate of 5e-5, decaying to a learning rate of 1e-5 using cosine scheduling. For AutoFormer-B, we fine-tune the supernet for 10 epochs with an initial learning rate of 2e-7, decaying to a learning rate of 1e-7 using cosine scheduling. All other settings, including batch size and data augmentation strategies, remain the same as in the training phase of GrowTAS.

\paragraph{Transfer Learning.}
We follow the standard transfer learning protocol~\cite{chen2021autoformer,wang2023prenas} to evaluate the generalization ability of the searched subnets. Specifically, for CIFAR-10, CIFAR-100~\cite{krizhevsky2009learning}, Flowers~\cite{nilsback2008automated}, and Cars~\cite{krause2013d}, we fine-tune the subnet using stochastic gradient descent (SGD) for 1000 epochs with an initial learning rate of 0.01 and a weight decay of 1e-4. For INAT-19~\cite{van2018inaturalist}, we use the AdamW optimizer~\cite{loshchilov2019decoupled} and fine-tune the subnet for 360 epochs with an initial learning rate of 7.5e-5 and a weight decay of 5e-2.

\begin{table}[t]
 \centering
 \small
 \renewcommand{\arraystretch}{1.2}
 \caption{Search space configurations for each AutoFormer space. Each tuple denotes (min, max, step) for the corresponding attribute.}
 \resizebox{\linewidth}{!}{%
 \begin{tabular}{lccc}
 \toprule
 
 Setting & Autoformer-T & Autoformer-S & Autoformer-B \\
 \midrule
 Embed Dim   & (192, 240, 24)   & (320, 448, 64)   & (528, 624, 48)   \\
 MLP Ratio   & (3.5, 4, 0.5)    & (3, 4, 0.5)      & (3, 4, 0.5)      \\
 Head Num    & (3, 4, 1)        & (5, 7, 1)        & (8, 10, 1)       \\
 Depth Num   & (12, 14, 1)      & (12, 14, 1)      & (14, 16, 1)      \\
 \midrule
 Params Range & 4--9M            & 14--34M          & 42--75M          \\
 \bottomrule
 \end{tabular}%
 }
 \label{tab:searchspace}
\end{table}

\subsection{Results} \label{sec:results}

\begin{table}[t]
  \centering
  \caption{
    Quantitative results of different ViT models on ImageNet~\cite{deng2009imagenet}. 
    We report the top-1 and top-5 validation accuracy (mean $\pm$ std over 3 seeds where available), 
    number of parameters, and FLOPs. 
    *: Results reproduced by us. 
    †: Results obtained using the training setting from PreNAS-T~\cite{wang2023prenas}.
  }
  \label{tab:vit_results}
  \renewcommand{\arraystretch}{1.2} 
  \normalsize
  \begin{adjustbox}{width=\linewidth}
  \begin{tabular}{lccccc}
  \toprule
  \textbf{Model} & \shortstack{\textbf{Top-1} \\ \textbf{(\%)}} & \shortstack{\textbf{Top-5} \\ \textbf{(\%)}} & \shortstack{\textbf{\#Params} \\ \textbf{(M)}} & \shortstack{\textbf{FLOPs} \\ \textbf{(G)}} \\
  \midrule
  ViT-T~\cite{dosovitskiy2020image}         & 74.5 & -    & 5.7  & -   \\
  DeiT-T~\cite{touvron2021training}         & 72.2 & 91.1 & 5.7  & 1.2 \\
  ConViT-T~\cite{dascoli2021convit}         & 73.1 & 91.7 & 5.7  & 1.2 \\
  TNT-T~\cite{han2021tnt}                   & 73.9 & 91.9 & 6.1  & 1.4 \\
  ViTAS-C~\cite{su2022vitas}                & 74.7 & -    & 5.8  & 1.2 \\
  FocusFormer-T~\cite{liu2022focusformer}   & 75.1 & 92.8 & 6.2  & 1.4 \\
  PreNAS-T†~\cite{wang2023prenas}           & 77.1 & 93.4 & 5.9  & 1.3 \\
  AutoFormer-T~\cite{chen2021autoformer}    & 74.7 & 92.6 & 5.9  & 1.3 \\
  AutoFormer-T*~\cite{chen2021autoformer}   & 74.9 & 92.5 & 5.9  & 1.3 \\
  DYNAS~\cite{jeon2025subnet}                 & 74.8 & -    & 5.9  & 1.4 \\
  \rowcolor{oursgray} \textbf{GrowTAS-T   } & 75.19 $\pm$ 0.01 & 92.60 $\pm$ 0.01 & 5.9  & 1.3 \\
  \rowcolor{oursgray} \textbf{GrowTAS-T+  } & 75.19 $\pm$ 0.01 & 92.60 $\pm$ 0.01 & 5.9  & 1.3 \\
  \rowcolor{oursgray} \textbf{GrowTAS-T† }  & 77.06 $\pm$ 0.00 & 93.37 $\pm$ 0.02 & 5.9  & 1.3 \\
  \midrule
  ViT-S~\cite{dosovitskiy2020image}         & 78.8 & -    & 22.1 & 5.1 \\
  DeiT-S~\cite{touvron2021training}         & 79.8 & 95.0 & 22.1 & 4.7 \\
  ConViT-S~\cite{dascoli2021convit}         & 81.3 & 95.7 & 27.0 & 5.4 \\
  TNT-S~\cite{han2021tnt}                   & 81.5 & 95.7 & 23.8 & 5.2 \\
  ViTAS-F~\cite{su2022vitas}                & 80.5 & -    & 27.6 & 5.5 \\
  FocusFormer-S~\cite{liu2022focusformer}   & 81.6 & -    & 23.7 & 5.1 \\
  PreNAS-S†~\cite{wang2023prenas}           & 81.8 & 95.9 & 22.9 & 5.1 \\
  AutoFormer-S~\cite{chen2021autoformer}    & 81.7 & 95.7 & 22.9 & 5.1 \\
  AutoFormer-S*~\cite{chen2021autoformer}   & 81.6 & 95.6 & 22.9 & 5.1 \\
  \rowcolor{oursgray} \textbf{GrowTAS-S  }  & 81.59 $\pm$ 0.02 & 95.76 $\pm$ 0.02 & 22.9 & 5.1 \\
  \rowcolor{oursgray} \textbf{GrowTAS-S+ }  & 81.75 $\pm$ 0.01 & 95.86 $\pm$ 0.04 & 22.9 & 5.1 \\
  \midrule
  PVT-Large~\cite{wang2021pvt}              & 81.7 & -    & 61.0 & 9.8 \\
  DeiT-B~\cite{touvron2021training}         & 81.8 & 95.6 & 86.0 & 18.0 \\
  ViT-B~\cite{dosovitskiy2020image}         & 79.7 & -    & 86.0 & 18.0 \\
  ConViT-B~\cite{dascoli2021convit}         & 82.4 & 95.9 & 86.0 & 17.0 \\
  FocusFormer-B~\cite{liu2022focusformer}   & 81.9 & 95.6 & 52.8 & 11.0 \\
  PreNAS-B†~\cite{wang2023prenas}           & 82.6 & 96.0 & 54.0 & 11.0 \\
  AutoFormer-B~\cite{chen2021autoformer}    & 82.4 & 95.7 & 54.0 & 11.0 \\
  AutoFormer-B*~\cite{chen2021autoformer}   & 82.4 & 95.9 & 54.0 & 11.0 \\
  \rowcolor{oursgray} \textbf{GrowTAS-B  }  & 82.60 $\pm$ 0.02 & 96.00 $\pm$ 0.03 & 53.2 & 11.0 \\
  \rowcolor{oursgray} \textbf{GrowTAS-B+ }  & 82.67 $\pm$ 0.00 & 96.06 $\pm$ 0.02 & 53.3 & 11.0 \\
  \bottomrule
  \end{tabular}
  \end{adjustbox}
\end{table}
  
  \definecolor{GrayRow}{gray}{0.93}
  \newcommand{\grayrow}{\rowcolor{GrayRow}}
  
\begin{table*}[t]
    \centering
    \small
    \caption{Quantitative results of transfer learning across multiple datasets: CIFAR-10/100~\cite{krizhevsky2009learning}, Flowers~\cite{nilsback2008automated}, Cars~\cite{krause2013d}, and INAT-19~\cite{van2018inaturalist}. We report top-1 accuracy (\%) and the number of model parameters. For GrowTAS-S (ours), we report mean $\pm$ std across 3 runs.}
    \label{tab:transfer_learning_comparison}
    \renewcommand{\arraystretch}{1.1}
    \begin{tabular}{lcccccc}
      \toprule
      \textbf{Model} & \textbf{\#Params (M)} & \textbf{CIFAR-10} & \textbf{CIFAR-100} & \textbf{Flowers} & \textbf{Cars} & \textbf{iNat-19} \\
      \midrule
      ViT-B/16~\cite{dosovitskiy2020image}     & 86.0 & 98.1 & 87.1 & 89.5 & - & -    \\
      ViT-L/16~\cite{dosovitskiy2020image}     & 307.0 & 97.9 & 86.4 & 89.7 & - & -    \\
      DeiT-B~\cite{touvron2021training}        & 86.0 & 99.1 & 90.8 & 98.4 & 92.1 & 77.7 \\
      ViTAE-S~\cite{xu2021vitae}               & 24.0 & 98.8 & 90.8 & 97.8 & 91.4 & 76.0 \\
      DearKD-S~\cite{chen2022dearkd}           & 22.0 & 98.4 & 89.3 & 97.4 & 91.3 & -    \\
      PreNAS-S~\cite{wang2023prenas}           & 23.0 & 99.1 & 91.2 & 97.6 & 92.2 & 76.4 \\
      AutoFormer-S~\cite{chen2021autoformer}   & 22.9 & 98.9 & 89.6 & 97.9 & 92.1 & 77.4 \\
      \grayrow
      \textbf{GrowTAS-S (Ours)}                & 22.4 & \textbf{98.95 $\pm$ 0.01} & \textbf{91.04 $\pm$ 0.13} & \textbf{98.21 $\pm$ 0.09} & \textbf{92.38 $\pm$ 0.04} & \textbf{77.94 $\pm$ 0.12} \\
      \bottomrule
    \end{tabular}
\end{table*}

\begin{figure*}[t]
  \centering
  \begin{subfigure}[t]{0.32\textwidth}
      \centering
      \includegraphics[width=\linewidth]{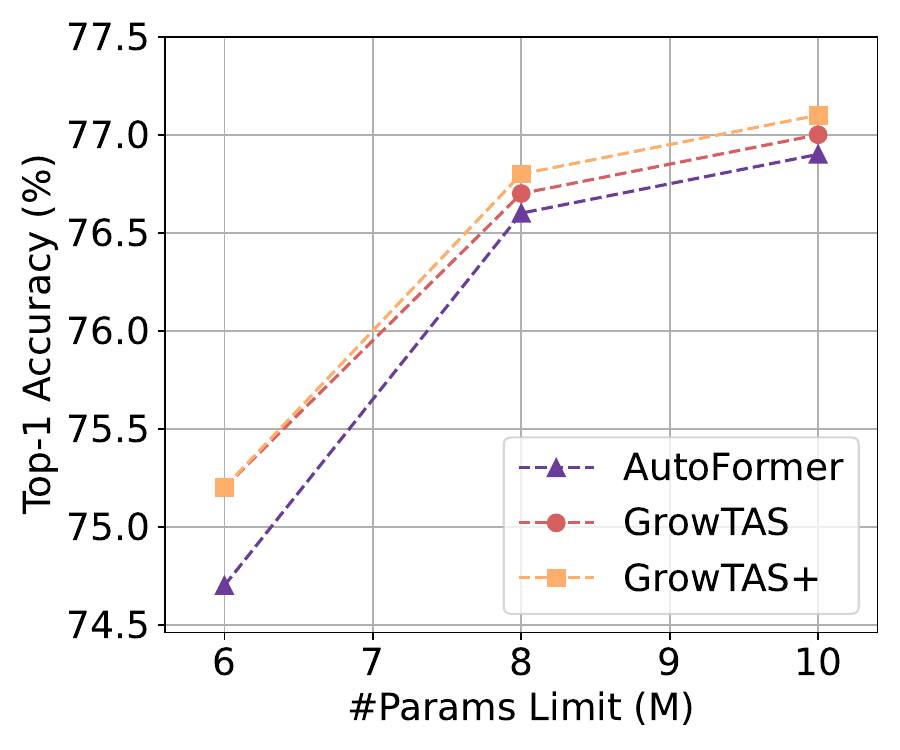}
      \caption{AutoFormer-T}
      \label{fig:autoformer_t}
  \end{subfigure}
  \hfill
  \begin{subfigure}[t]{0.32\textwidth}
      \centering
      \includegraphics[width=\linewidth]{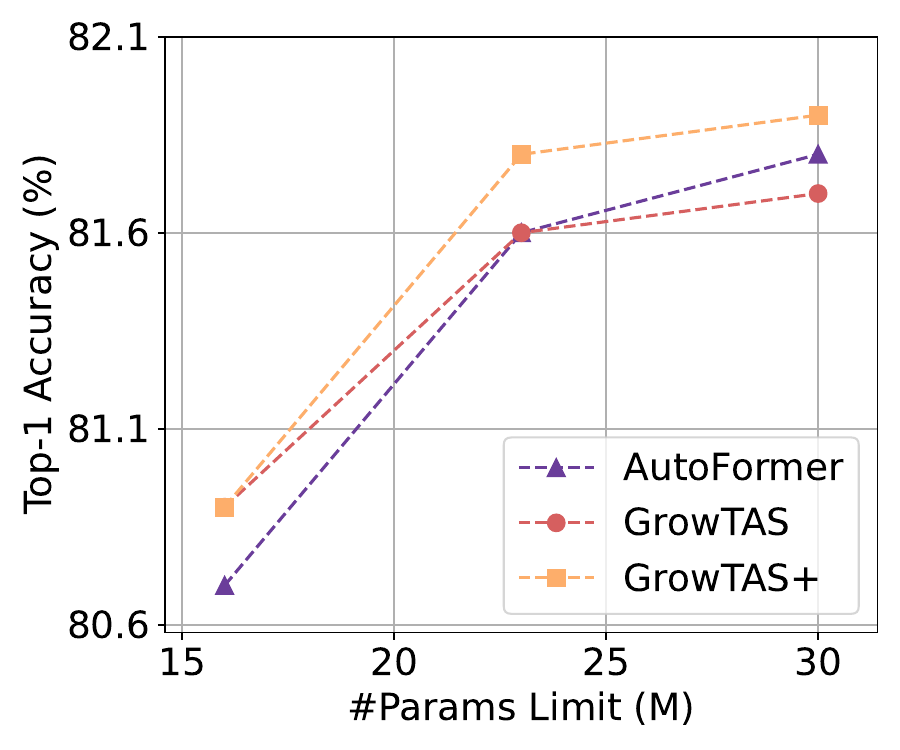}
      \caption{AutoFormer-S}
      \label{fig:autoformer_s}
  \end{subfigure}
  \hfill
  \begin{subfigure}[t]{0.32\textwidth}
      \centering
      \includegraphics[width=\linewidth]{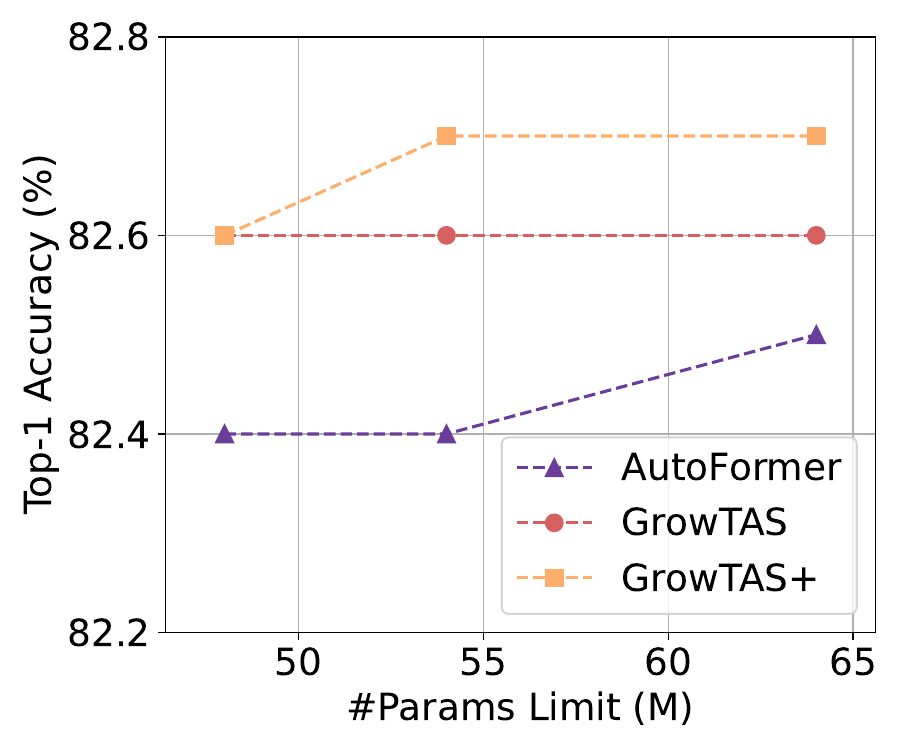}
      \caption{AutoFormer-B}
      \label{fig:autoformer_b}
  \end{subfigure}
  \caption{Top-1 accuracy on ImageNet~\cite{deng2009imagenet} under varying parameter limits in (a)~AutoFormer-T, (b)~AutoFormer-S, and (c)~AutoFormer-B search spaces.}
  \label{fig:accuracy_comparison}
\end{figure*}

\paragraph{ImageNet.} We show in Table~\ref{tab:vit_results} the comparisons of GrowTAS with state-of-the-art methods on ImageNet~\cite{deng2009imagenet}. We report the top-1 accuracy, the number of parameters, and FLOPs for each model. We observe the following key findings: (1) GrowTAS consistently outperforms the baseline AutoFormer~\cite{chen2021autoformer} across all spaces. This demonstrates the effectiveness of our progressive training strategy, where well-trained small subnets serve as good starting points for training larger ones. (2) GrowTAS achieves a better trade-off between accuracy and model complexity compared to others. For example, AutoFormer-B, PreNAS-B, and DeiT-B have parameter counts of 54.0M, 54.0M, and 86.0M, respectively, with top-1 accuracies in the range of 81.8--82.6\%. GrowTAS achieves a comparable or even higher accuracy of 82.6\% with fewer parameters of 53.2M, demonstrating its effectiveness. (3) GrowTAS+ further improves the performance of GrowTAS by fine-tuning only the subset of parameters introduced in the later stage of training. Specifically, subnets under the 6M constraint fall entirely within the smallest subspace $\mathcal{A}_1$, while the parameter constraints in GrowTAS-S and GrowTAS-B correspond to $\mathcal{A}_2$. As shown in the results, GrowTAS+ does not alter the performance of subnets from $\mathcal{A}_1$ subnets, but provides clear improvements for subnets from $\mathcal{A}_2$. This demonstrates that our fine-tuning strategy effectively enhances larger subnets without affecting the performance of smaller ones. Please refer to Fig.~\ref{fig:accuracy_comparison} for detailed analysis. (4) PreNAS-T outperforms GrowTAS-T. However, PreNAS requires tuning additional hyperparameters, such as data augmentation strategies and weight decay, specifically for the tiny search space. When we apply the same settings, our method, denoted by GrowTAS-T$\dagger$, achieves the performance comparable to PreNAS-T. Furthermore, PreNAS reduces the number of candidate subnets drastically (\eg, 6 out of \(2 \times 10^8\)) without training the remaining subnets.  It also requires re-selecting a few subnets and re-training them whenever hardware constraints are changed.

\paragraph{Transfer learning.}
We show in Table~\ref{tab:transfer_learning_comparison} the results of GrowTAS on various downstream classification tasks, including CIFAR-10/100~\cite{krizhevsky2009learning}, Flowers~\cite{nilsback2008automated}, Cars~\cite{krause2013d}, and iNat-19~\cite{van2018inaturalist}. We can see that GrowTAS-S achieves highly competitive or state-of-the-art performance across all benchmarks. These datasets span a broad spectrum of domains, covering general objects~\cite{krizhevsky2009learning}, fine-grained categories~\cite{nilsback2008automated, krause2013d}, and large-scale species classification~\cite{van2018inaturalist}. This suggests that our progressive subnet training strategy effectively preserves transferable features, demonstrating its generalization ability across diverse domains.


\begin{figure}[t]
  \centering
  \small
  \begin{subfigure}{1.0\linewidth}
    \centering
    \includegraphics[width=\linewidth]{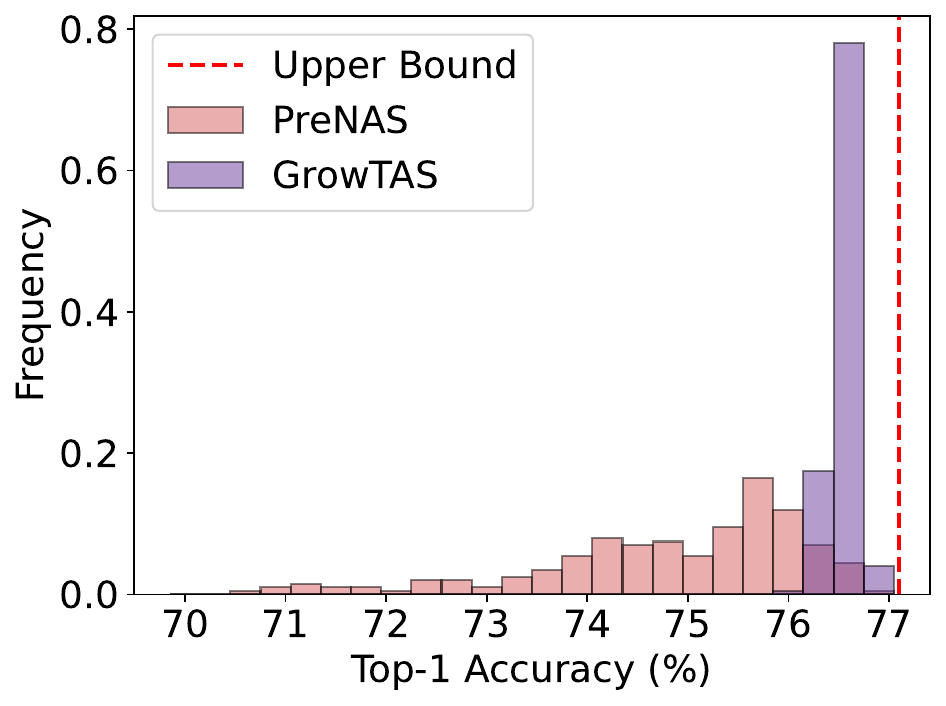}
  \end{subfigure}
  \caption{
    Top-1 accuracy distribution of 200 subnets sampled within the same constraint range ($\sim$6M). To ensure a fair comparison, GrowTAS is trained using the same settings as PreNAS. While both methods achieve the same upper bound of 77.1\%, GrowTAS exhibits significantly more subnets with high accuracy, while PreNAS has a narrow distribution with only a few subnets achieving high accuracy.}
  \label{fig:prenas_acc_dist}
\end{figure}

\definecolor{oursgray}{gray}{0.93}

\begin{table}[t]
  \centering
  \caption{Top-1 accuracies (\%) of AutoFormer, GrowTAS, and GrowTAS+ under varying transition steps \(T_1\) and parameter limits. Each model is evaluated at three parameter scales for AutoFormer-T/S/B search spaces~\cite{chen2021autoformer}.}
  \label{tab:growtas_results_all}
  \renewcommand{\arraystretch}{0.98} 
  \begin{adjustbox}{width=\linewidth}
    \begin{tabular}{lccccc}
      \toprule
      \multirow{2}{*}{Model} & \multirow{2}{*}{\#Params (M)} & \multicolumn{4}{c}{$T_1$}  \\
      \cmidrule(lr){3-6}
      & & 200 & 225 & 250 & 400 \\
      \midrule

    \multirow{3}{*}{AutoFormer-T} 
      & 6  & -- & -- & \cellcolor{oursgray}74.7 & -- \\
      & 8  & -- & -- & \cellcolor{oursgray}76.6 & -- \\
      & 10 & -- & -- & \cellcolor{oursgray}76.9 & -- \\
    \arrayrulecolor{lightgrayrule}\cmidrule(lr){1-6}\arrayrulecolor{black}

    \multirow{3}{*}{GrowTAS-T} 
      & 6  & 75.2 & 75.2 & \cellcolor{oursgray}75.2 & 75.3 \\
      & 8  & 76.7 & 76.6 & \cellcolor{oursgray}76.7 & 76.2 \\
      & 10 & 77.0 & 76.9 & \cellcolor{oursgray}77.0 & 76.3 \\
    \arrayrulecolor{lightgrayrule}\cmidrule(lr){1-6}\arrayrulecolor{black}

    \multirow{3}{*}{GrowTAS-T+} 
      & 6  & 75.2 & 75.2 & \cellcolor{oursgray}75.2 & 75.3 \\
      & 8  & 76.8 & 76.7 & \cellcolor{oursgray}76.8 & 76.7 \\
      & 10 & 77.1 & 77.0 & \cellcolor{oursgray}77.1 & 76.9 \\
    \midrule

    \multirow{3}{*}{AutoFormer-S} 
      & 16 & -- & \cellcolor{oursgray}80.7 & -- & -- \\
      & 23 & -- & \cellcolor{oursgray}81.6 & -- & -- \\
      & 30 & -- & \cellcolor{oursgray}81.8 & -- & -- \\
    \arrayrulecolor{lightgrayrule}\cmidrule(lr){1-6}\arrayrulecolor{black}

    \multirow{3}{*}{GrowTAS-S} 
      & 16 & 80.8 & \cellcolor{oursgray}80.9 & 80.7 & -- \\
      & 23 & 81.5 & \cellcolor{oursgray}81.6 & 81.6 & -- \\
      & 30 & 81.6 & \cellcolor{oursgray}81.7 & 81.7 & -- \\
    \arrayrulecolor{lightgrayrule}\cmidrule(lr){1-6}\arrayrulecolor{black}

    \multirow{3}{*}{GrowTAS-S+} 
      & 16 & 80.8 & \cellcolor{oursgray}80.9 & 80.7 & -- \\
      & 23 & 81.7 & \cellcolor{oursgray}81.8 & 81.7 & -- \\
      & 30 & 81.8 & \cellcolor{oursgray}81.9 & 81.8 & -- \\
    \midrule

    \multirow{3}{*}{AutoFormer-B} 
      & 48 & \cellcolor{oursgray}82.4 & -- & -- & -- \\
      & 54 & \cellcolor{oursgray}82.4 & -- & -- & -- \\
      & 64 & \cellcolor{oursgray}82.5 & -- & -- & -- \\
    \arrayrulecolor{lightgrayrule}\cmidrule(lr){1-6}\arrayrulecolor{black}

    \multirow{3}{*}{GrowTAS-B} 
      & 48 & \cellcolor{oursgray}82.6 & 82.4 & 82.6 & -- \\
      & 54 & \cellcolor{oursgray}82.6 & 82.4 & 82.6 & -- \\
      & 64 & \cellcolor{oursgray}82.6 & 82.4 & 82.5 & -- \\
    \arrayrulecolor{lightgrayrule}\cmidrule(lr){1-6}\arrayrulecolor{black}

    \multirow{3}{*}{GrowTAS-B+} 
      & 48 & \cellcolor{oursgray}82.6 & 82.4 & 82.6 & -- \\
      & 54 & \cellcolor{oursgray}82.7 & 82.5 & 82.6 & -- \\
      & 64 & \cellcolor{oursgray}82.7 & 82.5 & 82.6 & -- \\
    \bottomrule
  \end{tabular}
  \end{adjustbox}
\end{table}

\subsection{Analysis}
\label{sec:analysis}
\paragraph{Analysis under varying parameter constraints.}
We show in Fig.~\ref{fig:accuracy_comparison} the top-1 accuracies under varying parameter limits on ImageNet~\cite{deng2009imagenet} in AutoFormer-T, AutoFormer-S, and AutoFormer-B spaces. We can see that GrowTAS consistently outperforms the baseline AutoFormer across all parameter limits and search spaces. This suggests that our progressive subnet sampling strategy helps mitigate the weight conflict problem by gradually training subnets of different sizes. We can also see that the performance gap between GrowTAS and AutoFormer increases under lower parameter limits, indicating that the progressive subnet sampling strategy is particularly effective for smaller subnets, which have been especially susceptible to weight sharing. In addition, we observe that GrowTAS+ improves the performance of larger subnets without degrading the performance of smaller ones, demonstrating the effectiveness of fine-tuning strategy in GrowTAS+.


\paragraph{Ablations on transition step.}
The choice of the first transition step \(T_1\) plays a crucial role in balancing the training steps between small and large subnets. A higher \(T_1\) allows more training iterations for the subspace \(A_1\), which benefits small subnets to converge more effectively. On the other hand, a lower \(T_1\) provides more updates to subsequent larger search spaces, thereby improving the performance of larger subnets. To understand this trade-off, we show in Table~\ref{tab:growtas_results_all} the performance of GrowTAS under different choices of \(T_1\) across different parameter constraints. We can see that increasing \(T_1\) from 250 to 400 in GrowTAS-T slightly improves the performance of smaller subnets (75.2\% vs. 75.3\% at 6M), while reducing the performance of larger ones (77.1\% vs. 76.9\% at 10M). Given 500 total training epochs, we set \(T_1\) to the midpoint of 250 epochs and vary it around this value to find the best trade-off. We can see that the best trade-off between small and large subnet performance occurs at different \(T_1\) values depending on the model size, \ie,~250 for GrowTAS-T, 225 for GrowTAS-S, and 200 for GrowTAS-B. These values are fixed and used consistently throughout all experiments. It is important to note that while $T_1$ affects the balance between small and large subnets, GrowTAS achieves strong performance across all settings, as long as each subspace is allocated sufficient training time.

\paragraph{Comparison with PreNAS.} 
PreNAS~\cite{wang2023prenas} exploits a zero-cost proxy to select and train only a small subset of subnets. While efficient, this design limits generalization, where subnets not sampled during training remain unoptimized and consequently exhibit poor performance.  Consequently, if the target deployment scenario changes (\eg,~new hardware constraints), PreNAS must resample subnets and retrain the supernet from scratch to maintain performance. To validate this limitation, we sample 200 subnets randomly from both PreNAS and GrowTAS supernets and evaluate their performance. We show in Fig.~\ref{fig:prenas_acc_dist} the distribution of accuracies across these sampled subnets. We can see that GrowTAS maintains consistently high accuracy over a wide range of subnets, while PreNAS produces a significant number of low-performing subnets.  This suggests that GrowTAS is more robust to changes in deployment scenarios, as it can adapt to new constraints without requiring retraining. Notably, both methods achieve the same best accuracy of 77.1\% for the sampled subnets, indicating that GrowTAS matches the performance of PreNAS, while providing better performance across more subnets, demonstrating the effectiveness of GrowTAS.


\section{Limitations}

Our framework relies on a predefined transition step \(T\) to schedule the progressive expansion of the search space. This fixed schedule assumes that a static stage division is sufficient to balance the training of small and large subnets. While this approach is effective in practice, it may result in suboptimal stage allocation when applied to architectures or datasets with different convergence characteristics. A potential extension is to dynamically determine \(T\) by monitoring training signals such as loss reduction or convergence speed, which could enhance the adaptability and robustness of the framework across diverse scenarios.

\section{Conclusion}

We have tackled the weight interference problem in TAS, where shared weights hinder the performance of subnets, especially smaller ones. To address this, we have introduced GrowTAS, a progressive training framework that trains the supernet from small to large subnets to reduce interference and stabilize training. To improve the performance of large subnets, we have proposed GrowTAS+, which fine-tunes newly introduced weights for larger subnets while preserving the weights of smaller ones. Extensive experiments and analyses demonstrate the superiority of our approach over existing TAS methods. We believe that GrowTAS can provide a practical and scalable baseline for future research in TAS.

\paragraph{Acknowledgments}
This work was supported by Institute of Information \& Communications Technology Planning \& Evaluation (IITP) grants funded by the Korea government (MSIT) (No.RS-2022-00143524, Development of Fundamental Technology and Integrated Solution for Next-Generation Automatic Artificial Intelligence System, No.RS-2025-09942968, AI Semiconductor Innovation Lab (Yonsei University)), and the National Research Foundation of Korea (NRF) grants funded by the Korea government (MSIT) (RS-2025-02216328).

{
    \small
    \bibliographystyle{ieeenat_fullname}
    \bibliography{main}

@String(CVPR= {IEEE Conf. Comput. Vis. Pattern Recog.})

@String(ICCV= {Int. Conf. Comput. Vis.})

@String(ECCV= {Eur. Conf. Comput. Vis.})

@String(ICLR = {Int. Conf. Learn. Represent.})

@String(AAAI = {AAAI})

@String(CVPRW= {IEEE Conf. Comput. Vis. Pattern Recog. Worksh.})

@String(CVPR  = {CVPR})

@String(ICCV  = {ICCV})

@String(ECCV  = {ECCV})

@String(ICLR  = {ICLR})

@String(CVPRW= {CVPRW})

@inproceedings{cai2020once,
  title={Once-for-All: Train one network and specialize it for efficient deployment},
  author={Cai, Han and Gan, Chuang and Wang, Tianzhe and Zhang, Zhekai and Han, Song},
  booktitle={International Conference on Learning Representations (ICLR)},
  year={2020}
}

@inproceedings{chen2021autoformer,
  title={AutoFormer: Searching transformers for visual recognition},
  author={Chen, Minghao and Peng, Houwen and Fu, Jianlong and Ling, Haibin},
  booktitle={Proceedings of the IEEE/CVF International Conference on Computer Vision (ICCV)},
  pages={12270--12280},
  year={2021}
}

@inproceedings{he2016deep,
  title={Deep residual learning for image recognition},
  author={He, Kaiming and Zhang, Xiangyu and Ren, Shaoqing and Sun, Jian},
  booktitle={Proceedings of the IEEE Conference on Computer Vision and Pattern Recognition (CVPR)},
  pages={770--778},
  year={2016}
}

@inproceedings{tan2019efficientnet,
  title={EfficientNet: Rethinking model scaling for convolutional neural networks},
  author={Tan, Mingxing and Le, Quoc},
  booktitle={Proceedings of the International Conference on Machine Learning (ICML)},
  pages={6105--6114},
  year={2019}
}

@inproceedings{su2022vitas,
  title={VITAS: Vision Transformer Architecture Search},
  author={Su, Xiu and You, Shan and Xie, Jiyang and Zheng, Mingkai and Wang, Fei and Qian, Chen and Zhang, Changshui and Wang, Xiaogang and Xu, Chang},
  booktitle={European Conference on Computer Vision (ECCV)},
  year={2022}
}

@inproceedings{goldberg1991comparative,
  title={A comparative analysis of selection schemes used in genetic algorithms},
  author={Goldberg, David E and Deb, Kalyanmoy},
  booktitle={Foundations of Genetic Algorithms},
  pages={69--93},
  year={1991},
  publisher={Elsevier}
}

@inproceedings{yu2020bignas,
  title={BigNAS: Scaling Up Neural Architecture Search with Big Single-Stage Models},
  author={Yu, Jiahui and Yang, Linjie and Xu, Ning and Shen, Xiaodan and Huang, Thomas},
  booktitle={ECCV},
  pages={702--717},
  year={2020},
  organization={Springer}
}

@techreport{krizhevsky2009learning,
  title={Learning multiple layers of features from tiny images},
  author={Krizhevsky, Alex and Hinton, Geoffrey and others},
  year={2009},
  institution={University of Toronto}
}

@inproceedings{deng2009imagenet,
  title={ImageNet: A large-scale hierarchical image database},
  author={Deng, Jia and Dong, Wei and Socher, Richard and Li, Li-Jia and Li, Kai and Fei-Fei, Li},
  booktitle={IEEE Conference on Computer Vision and Pattern Recognition (CVPR)},
  pages={248--255},
  year={2009},
  organization={IEEE}
}

@inproceedings{nilsback2008automated,
  title={Automated flower classification over a large number of classes},
  author={Nilsback, Maria-Elena and Zisserman, Andrew},
  booktitle={Sixth Indian Conference on Computer Vision, Graphics \& Image Processing},
  year={2008},
  organization={IEEE}
}

@inproceedings{krause2013d,
  title={3D object representations for fine-grained categorization},
  author={Krause, Jonathan and Stark, Michael and Deng, Jia and Fei-Fei, Li},
  booktitle={IEEE International Conference on Computer Vision Workshops (ICCVW)},
  year={2013},
  organization={IEEE}
}

@inproceedings{van2018inaturalist,
  title={The iNaturalist species classification and detection dataset},
  author={Van Horn, Grant and Mac Aodha, Oisin and Song, Yang and Cui, Yin and Sun, Chen and Shepard, Alex and Adam, Hartwig and Perona, Pietro and Belongie, Serge},
  booktitle={IEEE Conference on Computer Vision and Pattern Recognition (CVPR)},
  year={2018},
  organization={IEEE}
}

@article{dosovitskiy2020image,
  title={An Image is Worth 16×16 Words: Transformers for Image Recognition at Scale},
  author={Dosovitskiy, Alexey and Beyer, Lucas and Kolesnikov, Alexander and Weissenborn, Dirk and Zhai, Xiaohua and Unterthiner, Thomas and Dehghani, Mostafa and Minderer, Matthias and Heigold, Georg and Gelly, Sylvain and Others},
  journal={arXiv preprint arXiv:2010.11929},
  year={2020}
}

@inproceedings{carion2020end,
  title={End-to-End Object Detection with Transformers},
  author={Carion, Nicolas and Massa, Francisco and Synnaeve, Gabriel and Usunier, Nicolas and Kirillov, Alexander and Zagoruyko, Sergey},
  booktitle={Proceedings of the IEEE/CVF Conference on Computer Vision and Pattern Recognition (CVPR)},
  pages={213--229},
  year={2020},
  organization={IEEE}
}

@inproceedings{strudel2021segmenter,
  title={Segmenter: Transformer for Semantic Segmentation},
  author={Strudel, Richard and Garcia, Marco and Laptev, Ivan and Schmid, Cordelia},
  booktitle={Proceedings of the IEEE/CVF International Conference on Computer Vision (ICCV)},
  pages={11125--11136},
  year={2021},
  organization={IEEE}
}

@inproceedings{lee2024aznas,
  title={AZ-NAS: Assembling Zero-Cost Proxies for Network Architecture Search},
  author={Lee, Junghyup and Ham, Bumsub},
  booktitle={Proceedings of the IEEE/CVF Conference on Computer Vision and Pattern Recognition (CVPR)},
  year={2024}
}

@article{liu2022focusformer,
  author    = {Jing Liu and Jianfei Cai and Bohan Zhuang},
  title     = {Focusformer: Focusing on what we need via architecture sampler},
  journal   = {arXiv preprint arXiv:2208.10861},
  year      = {2022}
}

@inproceedings{zhang2023shiftnas,
  author    = {Mingyang Zhang and Xinyi Yu and Haodong Zhao and Linlin Ou},
  title     = {Shiftnas: Improving one-shot nas via probability shift},
  booktitle = {Proceedings of the IEEE/CVF International Conference on Computer Vision (ICCV)},
  year      = {2023}
}

@inproceedings{wang2023prenas,
  author    = {Haibin Wang and Ce Ge and Hesen Chen and Xiuyu Sun},
  title     = {Prenas: Preferred one-shot learning towards efficient neural architecture search},
  booktitle = {Proceedings of the 40th International Conference on Machine Learning (ICML)},
  year      = {2023}
}

@inproceedings{guo2020spos,
  author    = {Guo, Zichao and Zhang, Xiangyu and Mu, Haoyuan and Heng, Wen and Liu, Zechun and Wei, Yichen and Sun, Jian},
  title     = {Single Path One-Shot Neural Architecture Search with Uniform Sampling},
  booktitle = {European Conference on Computer Vision (ECCV)},
  year      = {2020},
  pages     = {544--560},
  publisher = {Springer},
  doi       = {10.1007/978-3-030-58592-1_33}
}

@inproceedings{you2020greedynas,
  author    = {Shangqian You and Tao Huang and Mingmin Yang and Fei Wang and Chen Qian and Changxin Zhang},
  title     = {GreedyNAS: Towards Fast One-Shot NAS with Greedy Supernet},
  booktitle = {IEEE/CVF Conference on Computer Vision and Pattern Recognition (CVPR)},
  year      = {2020},
  pages     = {1999--2008},
  publisher = {IEEE},
  doi       = {10.1109/CVPR42600.2020.00207}
}

@inproceedings{lee2019snip,
  title={SNIP: Single-shot network pruning based on connection sensitivity},
  author={Lee, Namhoon and Ajanthan, Theja and Torr, Philip HS},
  booktitle={International Conference on Learning Representations (ICLR)},
  year={2019}
}

@inproceedings{vaswani2017attention,
  title={Attention is All You Need},
  author={Vaswani, Ashish and Shazeer, Noam and Parmar, Niki and Uszkoreit, Jakob and Jones, Llion and Gomez, Aidan N and Kaiser, Łukasz and Polosukhin, Illia},
  booktitle={Advances in Neural Information Processing Systems (NeurIPS)},
  year={2017}
}

@inproceedings{guo2020single,
  title={Single Path One-Shot Neural Architecture Search with Uniform Sampling},
  author={Guo, Zichao and Zhang, Xiangyu and Mu, Haoyuan and Heng, Wen and Liu, Zechun and Wei, Yichen and Sun, Jian},
  booktitle={European Conference on Computer Vision (ECCV)},
  year={2020}
}

@inproceedings{pham2018efficient,
  title={Efficient Neural Architecture Search via Parameter Sharing},
  author={Pham, Hieu and Guan, Melody Y and Zoph, Barret and Le, Quoc V and Dean, Jeff},
  booktitle={International Conference on Machine Learning (ICML)},
  year={2018}
}

@article{williams1992simple,
  author    = {Williams, Ronald J.},
  title     = {Simple statistical gradient-following algorithms for connectionist reinforcement learning},
  journal   = {Machine learning},
  year      = {1992}
}

@inproceedings{zoph2017nas,
  author    = {Zoph, Barret and Le, Quoc V.},
  title     = {Neural architecture search with reinforcement learning},
  booktitle = {International Conference on Learning Representations (ICLR)},
  year      = {2017}
}

@inproceedings{zoph2018learning,
  author    = {Zoph, Barret and Vasudevan, Vijay and Shlens, Jonathon and Le, Quoc V.},
  title     = {Learning transferable architectures for scalable image recognition},
  booktitle = {Proceedings of the IEEE Conference on Computer Vision and Pattern Recognition (CVPR)},
  year      = {2018}
}

@inproceedings{baker2017designing,
  author    = {Baker, Bowen and Gupta, Otkrist and Naik, Nikhil and Raskar, Ramesh},
  title     = {Designing neural network architectures using reinforcement learning},
  booktitle = {International Conference on Learning Representations (ICLR)},
  year      = {2017}
}

@inproceedings{yu2019slimmable,
  author    = {Yu, Jiahui and Yang, Lin and Xu, Ning and Yang, Jianchao and Huang, Thomas},
  title     = {Slimmable neural networks},
  booktitle = {International Conference on Learning Representations (ICLR)},
  year      = {2019}
}

@inproceedings{touvron2021training,
  title     = {Training data-efficient image transformers \& distillation through attention},
  author    = {Touvron, Hugo and Cord, Matthieu and Douze, Matthijs and Massa, Francisco and Sablayrolles, Alexandre and Jégou, Hervé},
  booktitle = {Proceedings of the 38th International Conference on Machine Learning (ICML)},
  year      = {2021}
}

@inproceedings{dascoli2021convit,
  title     = {ConViT: Improving Vision Transformers with Soft Convolutional Inductive Biases},
  author    = {d’Ascoli, Stéphane and Touvron, Hugo and Leavitt, Matthew L and Morcos, Ari S and Biroli, Giulio and Sagun, Levent},
  booktitle = {Proceedings of the 38th International Conference on Machine Learning (ICML)},
  year      = {2021}
}

@inproceedings{han2021tnt,
  title     = {Transformer in Transformer},
  author    = {Han, Kai and Xiao, An and Wu, Enhua and Guo, Jianyuan and Xu, Chunjing and Wang, Yunhe},
  booktitle = {Advances in Neural Information Processing Systems (NeurIPS)},
  year      = {2021}
}

@inproceedings{wang2021pvt,
  title     = {Pyramid Vision Transformer: A Versatile Backbone for Dense Prediction without Convolutions},
  author    = {Wang, Wenhai and Xie, Enze and Li, Xiang and Fan, Deng-Ping and Song, Kaitao and Liang, Ding and Lu, Tong and Luo, Ping and Shao, Ling},
  booktitle = {Proceedings of the IEEE/CVF International Conference on Computer Vision (ICCV)},
  pages     = {568--578},
  year      = {2021}
}

@inproceedings{paszke2019pytorch,
  title     = {PyTorch: An Imperative Style, High-Performance Deep Learning Library},
  author    = {Paszke, Adam and Gross, Sam and Massa, Francisco and Lerer, Adam and Bradbury, James and Chanan, Gregory and Killeen, Trevor and Lin, Zeming and Gimelshein, Natalia and Antiga, Luca and Desmaison, Alban and Kopf, Andreas and Yang, Edward and DeVito, Zachary and Raison, Martin and Tejani, Alykhan and Chilamkurthy, Sasank and Steiner, Benoit and Fang, Lu and Bai, Junjie and Chintala, Soumith},
  booktitle = {Advances in Neural Information Processing Systems},
  volume    = {32},
  pages     = {8024--8035},
  year      = {2019}
}

@misc{wightman2019timm,
  author       = {Wightman, Ross},
  title        = {PyTorch Image Models},
  year         = {2019},
  howpublished = {\url{https://github.com/rwightman/pytorch-image-models}},
}

@inproceedings{xu2021vitae,
  title     = {ViTAE: Vision Transformer Advanced by Exploring Intrinsic Inductive Bias},
  author    = {Xu, Yufei and Zhang, Qiming and Zhang, Jing and Tao, Dacheng},
  booktitle = {Advances in Neural Information Processing Systems (NeurIPS)},
  year      = {2021}
}

@inproceedings{chen2022dearkd,
  title     = {DearKD: Data-Efficient Early Knowledge Distillation for Vision Transformers},
  author    = {Chen, Xianing and Cao, Qiong and Zhong, Yujie and Zhang, Jing and Gao, Shenghua and Tao, Dacheng},
  booktitle = {Proceedings of the IEEE/CVF Conference on Computer Vision and Pattern Recognition (CVPR)},
  year      = {2022}
}

@inproceedings{oh2025efficient,
  title     = {Efficient Few-Shot Neural Architecture Search by Counting the Number of Nonlinear Functions},
  author    = {Youngmin Oh and Hyunju Lee and Bumsub Ham},
  booktitle = {Proceedings of the AAAI Conference on Artificial Intelligence},
  year      = {2025}
}

@inproceedings{jeon2025subnet,
  title     = {Subnet-Aware Dynamic Supernet Training for Neural Architecture Search},
  author    = {Jeimin Jeon and Youngmin Oh and Junghyup Lee and Donghyeon Baek and Dohyung Kim and Chanho Eom and Bumsub Ham},
  booktitle = {Proceedings of the IEEE/CVF Conference on Computer Vision and Pattern Recognition (CVPR)},
  pages     = {30137--30146},
  year      = {2025}
}

@inproceedings{loshchilov2019decoupled,
  title     = {Decoupled Weight Decay Regularization},
  author    = {Loshchilov, Ilya and Hutter, Frank},
  booktitle = {International Conference on Learning Representations (ICLR)},
  year      = {2019},
  url       = {https://openreview.net/forum?id=Bkg6RiCqY7}
}

@inproceedings{cubuk2020randaugment,
  author    = {Ekin D. Cubuk and Barret Zoph and Jonathon Shlens and Quoc V. Le},
  title     = {RandAugment: Practical automated data augmentation with a reduced search space},
  booktitle = {Proc. IEEE/CVF Conf. Comput. Vis. Pattern Recognit. Workshops (CVPRW)},
  pages     = {3008--3017},
  year      = {2020}
}

@inproceedings{zhang2018mixup,
  author    = {Hongyi Zhang and Moustapha Cisse and Yann N. Dauphin and David Lopez-Paz},
  title     = {mixup: Beyond Empirical Risk Minimization},
  booktitle = {Proc. Int. Conf. Learn. Represent. (ICLR)},
  year      = {2018}
}

@inproceedings{szegedy2016rethinking,
  author    = {Christian Szegedy and Vincent Vanhoucke and Sergey Ioffe and Jonathon Shlens and Zbigniew Wojna},
  title     = {Rethinking the Inception Architecture for Computer Vision},
  booktitle = {Proc. IEEE Conf. Comput. Vis. Pattern Recognit. (CVPR)},
  pages     = {2818--2826},
  year      = {2016}
}

@inproceedings{yuan2020revisiting,
  author    = {Li Yuan and Francis E. H. Tay and Guilin Li and Tao Wang and Jiashi Feng},
  title     = {Revisiting Knowledge Distillation via Label Smoothing Regularization},
  booktitle = {Proc. IEEE/CVF Conf. Comput. Vis. Pattern Recognit. (CVPR)},
  pages     = {3902--3910},
  year      = {2020}
}

@article{berman2019multigrain,
  author    = {Maxim Berman and Herv\'e J\'egou and Andrea Vedaldi and Iasonas Kokkinos and Matthijs Douze},
  title     = {MultiGrain: A unified image embedding for classes and instances},
  journal   = {arXiv preprint arXiv:1902.05509},
  year      = {2019}
}

@inproceedings{hoffer2020augment,
  author    = {Elad Hoffer and Tal Ben-Nun and Itay Hubara and Nir Giladi and Torsten Hoefler and Daniel Soudry},
  title     = {Augment Your Batch: Improving Generalization through Instance Repetition},
  booktitle = {Proc. IEEE/CVF Conf. Comput. Vis. Pattern Recognit. (CVPR)},
  pages     = {8126--8135},
  year      = {2020}
}

@inproceedings{yun2019cutmix,
  author    = {Sangdoo Yun and Dongyoon Han and Seong Joon Chun and Sanghyuk Johan Oh and Youngjoon Yoo and Junsuk Choe},
  title     = {CutMix: Regularization Strategy to Train Strong Classifiers with Localizable Features},
  booktitle = {Proc. IEEE/CVF Int. Conf. Comput. Vis. (ICCV)},
  pages     = {6022--6031},
  year      = {2019}
}
}


\end{document}